\documentclass[letterpaper]{article}
\usepackage{amsmath}
\usepackage{aaai}
\usepackage{times}
\usepackage{named}
\usepackage{helvet}
\usepackage{courier}

\def\ar{\leftarrow}
\def\beq{\begin{equation}}
\def\eeq#1{\label{#1}\end{equation}}
\def\ba{\begin{array}}
\def\ea{\end{array}}
\def\i#1{\hbox{\it #1\/}}

\def\ar{\leftarrow}
\def\rar{\rightarrow}
\def\lrar{\leftrightarrow}
\def\no{\i{not}}

\def\i#1{\hbox{\itshape #1\/}}

\newtheorem{prop}{Proposition}

\newtheorem{cor}{Corollary}
\newtheorem{lemma}{Lemma}

\newtheorem{definition}{Definition}

\newtheorem{example}{Example}

\def\snes{\hbox{\i{NES}}}
\def\fnes{\hbox{\i{NFES}}}
\def\NE{\hbox{\i{Nonempty}}}
\def\flf{\hbox{\i{FLF}}}
\def\loop{\hbox{\i{Loop}}}
\def\fes{\hbox{\i{FES}}}
\def\oa{\overrightarrow}
\def\smmodels{\models_{\text{\sm}}}

\newcommand{\myar}{\hspace{-0.8em}\ar\hspace{-0.8em}}

\def\circ{\hbox{\rm CIRC}}
\def\sm{\hbox{\rm SM}}

\title{On Loop Formulas with Variables}

\author{
Joohyung Lee and Yunsong Meng\\
Computer Science and Engineering \\
School of Computing and Informatics \\
Arizona State University, Tempe, USA \\
{\tt \{joolee, Yunsong.Meng\}@asu.edu}
}

\begin{document}

\maketitle

\begin{abstract}
Recently Ferraris, Lee and Lifschitz proposed a new definition of
stable models that does not refer to grounding, which applies to the
syntax of arbitrary first-order sentences. We show its relation to the
idea of loop formulas with variables by Chen, Lin, Wang and Zhang, and
generalize their loop formulas to disjunctive programs and to arbitrary
first-order sentences. We also extend the syntax of logic programs
to allow explicit quantifiers, and define its semantics as a subclass of
the new language of stable models by Ferraris {\sl et al.} Such programs
inherit from the general language the ability to handle nonmonotonic 
reasoning under the stable model semantics even in the absence of the
unique name and the domain closure assumptions, while yielding more
succinct loop formulas than the general language due to the restricted 
syntax. We also show certain syntactic conditions under which query
answering for an extended program can be reduced to entailment
checking in first-order logic, providing a way to apply 
first-order theorem provers to reasoning about non-Herbrand stable models.
\end{abstract}

\section{Introduction}\label{sec:intro}

The theorem on loop formulas showed that the stable models (answer sets) 
are the models of the logic program that satisfy all its loop
formulas. This idea has turned out to be widely applicable in relating
the stable model semantics~\cite{gel88} to propositional logic, which in
turn allowed to use SAT solvers for computing answer sets.  Since the
original invention of loop formulas for nondisjunctive logic
programs~\cite{lin04}, the theorem has been extended to more
general classes of logic programs, such as disjunctive
programs~\cite{lee03a}, programs with classical negation and infinite
programs~\cite{lee05}, arbitrary propositional formulas under the
stable model semantics~\cite{fer06}, and programs with
aggregates~\cite{liu05a}. The theorem has also been applied to other
nonmonotonic formalisms, such as nonmonotonic causal
theories~\cite{lee03b} and McCarthy's circumscription~\cite{lee06}.
The notion of a loop has been further refined by ``elementary
sets''~\cite{geb06}.

However, most work has been restricted to the propositional
case. Variables contained in a program are first eliminated by 
grounding---the process which replaces every variable with 
every object constant---and then loop formulas are computed from the
ground program. As a result, loop formulas were defined as formulas in 
propositional logic. 

\citeauthor{che06}'s definition of loop formulas [\citeyear{che06}]
is different in that loop formulas are obtained from the original
program without converting to the ground program, so that variables
remain. However, since the underlying semantics of logic programs
refers to grounding, such a loop formula was understood as a schema 
for the set of propositional loop formulas.

Recently there emerged a generalization of the stable model semantics
that does not refer to grounding~\cite{fer07a}. The semantics turns a
first-order sentence into a second-order sentence using the ``stable
model operator'' {\sm}, similar to the use of the ``circumscription
operator'' {\circ}~\cite{mcc80}. Logic programs are understood as a
special class of first-order sentences under the stable model
semantics. Unlike the traditional stable model semantics, the new
language has quantifiers with genuine object variables and the
notion of first-order models instead of Herbrand models. Consequently,
as in classical logic, it has no built-in unique name and domain
closure assumptions.

In this paper, we study the relationship between first-order loop
formulas from~\cite{che06} and the new definition of stable models
from~\cite{fer07a}. We also extend the definition of first-order loop
formulas from \cite{che06} to disjunctive programs and to arbitrary
first-order sentences, and present certain conditions under which the
new second-order definition of stable models can be turned into 
formulas in first-order logic in the form of loop formulas.

The studied relationship helps extend the syntax of logic programs
by allowing explicit quantifiers, which will be useful in overcoming
the difficulties of traditional answer set programs in reasoning about the
existence (or absence) of unnamed objects. 
We define the semantics of extended programs as a subclass of the new
language of stable models from \cite{fer07a}. Such programs inherit
from the general language the ability to handle nonmonotonic reasoning
under the stable model semantics even in the absence of the unique
name and the domain closure assumptions. On the other hand, extended
programs yield succinct loop formulas due to the restricted syntax so
that it is feasible to apply first-order theorem provers as
computational engines.

Imagine an insurance policy considering ``a person is eligible for a discount
plan if he or she has a spouse and has no record of accident.'' This can be
represented by the following program~$\Pi_1$ that contains explicit 
existential quantifiers.
\[
\ba {rcl}
  \i{GotMarried}(x,y)&\myar&\i{Spouse}(x,y) \\
  \i{Spouse}(x,y)&\myar&\i{GotMarried}(x,y), \no\ \i{Divorced}(x,y) \\
  \exists w\ \i{Discount}(x,w)&\myar&
        \i{Spouse}(x,y), \no\ \exists z\ \i{Accident}(x,z)
\ea
\]
We will say that a program~$\Pi$ entails a query $F$ (under the stable
model semantics) if every stable model of $\Pi$ satisfies~$F$. For example,
\begin{itemize}
\item  $\Pi_1$ entails each of $\neg\exists xy\ \i{Spouse}(x,y)$ and
       $\neg\exists xy\ \i{Discount}(x,y)$.

\item  $\Pi_1$ conjoined with $\Pi_2 = \{\exists y\
        \i{GotMarried}(\i{marge},y)\}$,
       no more entails $\neg\exists xw\ \i{Discount}(x,w)$, but
       entails each of $\exists xw\ \i{Discount}(x,w)$ and
       \hbox{$\forall x (\i{Discount}(x,\i{plan1})\rar x=\i{marge})$}.

\item  $\Pi_1$ conjoined with
       \[ \Pi_3 = \{\i{Spouse}(\i{homer},\i{marge}),
                  \exists z\ \i{Accident}(\i{homer},z)\}
        \]
        entails
        $\neg \exists w\ \i{Discount}(\i{homer},w)$.
\end{itemize}

For the reasoning of this kind, we need the notion of non-Herbrand
models since the names of discount plans, spouses and accident records
may be unknown. However, answer sets are defined as a special class of
Herbrand models.  Instead, we will show how reasoning about
non-Herbrand stable models can be represented by extended programs
and can be computed using loop formulas with variables. This provides
a way to apply first-order theorem provers to reasoning about
non-Herbrand stable models.

The paper is organized as follows. In the next section we review the
new definition of stable models from~\cite{fer07a}. Then we
review first-order loop formulas from~\cite{che06} and extend the result
to disjunctive programs and to arbitrary sentences. We compare the new
definition of stable models with first-order loop formulas and show
certain conditions under which the former can be reduced to the latter.
Given these results we give the notion of extended programs with
explicit quantifiers and show how query answering for extended programs can be
reduced to entailment checking in first-order logic.

\section{Review of the New Stable Model Semantics}

Let {\bf p} be a list of distinct predicate constants $p_1,\dots,p_n$,
and let {\bf u} be a list of distinct predicate variables
$u_1,\dots,u_n$ of the same length as {\bf p}.  By ${\bf u}={\bf p}$
we denote the conjunction of the formulas
$\forall {\bf x}(u_i({\bf x})\lrar p_i({\bf x}))$, where {\bf x} is a list
of distinct object variables of the same arity as the length of $p_i$, for
all $i=1,\dots n$.  By ${\bf u}\leq{\bf p}$ we denote the conjunction of the
formulas $\forall {\bf x}(u_i({\bf x})\rar p_i({\bf x}))$ for all
$i=1,\dots n$, and ${\bf u}<{\bf p}$ stands for
$({\bf u}\leq{\bf p})\land\neg({\bf u}={\bf p})$.

For any first-order sentence $F$, $\sm[F]$ stands for the second-order sentence
$$
   F \land \neg \exists {\bf u} (({\bf u}<{\bf p}) \land F^*({\bf u})),
$$
where {\bf p} is the list $p_1,\dots,p_n$ of all predicate constants occurring
in $F$, {\bf u} is a list $u_1,\dots,u_n$ of distinct predicate variables
of the same length as {\bf p}, and $F^*({\bf u})$ is defined recursively:
\begin{itemize}
\item  $p_i(t_1,\dots,t_m)^* = u_i(t_1,\dots,t_m)$;
\item  $(t_1\!=\!t_2)^* = (t_1\!=\!t_2)$;
\item  $\bot^* = \bot$;
\item  $(F\land G)^* = F^* \land G^*$;
\item  $(F\lor G)^* = F^* \lor G^*$;
\item  $(F\rar G)^* = (F^* \rar G^*)\land (F \rar G)$;
\item  $(\forall xF)^* = \forall xF^*$;
\item  $(\exists xF)^* = \exists xF^*$.
\end{itemize}
(There is no clause for negation here, because we treat $\neg F$ as shorthand
for $F\rar\bot$.)  According to~\cite{fer07a}, an interpretation of the
signature~$\sigma(F)$ consisting of the object, function and predicate
constants occurring in~$F$ is a {\sl stable model} of~$F$ if it satisfies
$\sm[F]$.

The terms ``stable model'' and ``answer set'' are often used in the literature
interchangeably. In the context of the new language of stable models,
it is convenient to distinguish between them as follows:
By an {\sl answer set}
of a first-order sentence~$F$ that contains at least one object constant we
will understand an Herbrand\footnote{Recall that an {\sl Herbrand
interpretation} of a signature~$\sigma$ (containing at least one object
constant) is an interpretation of~$\sigma$ such that its universe is the
set of all ground terms of~$\sigma$, and every ground term represents
itself.  An Herbrand interpretation can be identified with the set of ground
atoms to which it assigns the value {\sl true}.} interpretation of~$\sigma(F)$
that satisfies~$\sm[F]$.

Logic programs are viewed as alternative notation for first-order
sentences of special kinds (called the FOL-representation) by
\begin{itemize}
\item
replacing every comma by $\land$, every semi-colon by $\lor$, and
every $\no$ by $\neg\,$;
\item
turning every rule $\i{Head}\ar\i{Body}$ into a formula by rewriting it as
the implication $\i{Body}\rar\i{Head}$, and
\item
forming the conjunction of the universal closures of these formulas.
\end{itemize}

\begin{example}
For program $\Pi$ that contains three rules
\[
\ba l
  p(a)\\
  q(b)\\
  r(x)\ar p(x), \no\ q(x)
\ea
\]
the FOL-representation $F$ of $\Pi$ is
\beq
  p(a)\land q(b)\land\forall x((p(x)\land\neg q(x))\rar r(x))
\eeq{ex3f}
and $\sm[F]$ is 
$$
\ba l
p(a)\land q(b)\land\forall x((p(x)\land\neg q(x))\rar r(x))\\
\quad\land\neg\exists uvw(((u,v,w)<(p,q,r))\wedge u(a)\land v(b)\\
\qquad\qquad\qquad\land\forall x(((u(x)\land
(\neg v(x)\land\neg q(x))
)\rar w(x))\\
\qquad\qquad\qquad\quad\;\;\land((p(x)\land\neg q(x))\rar r(x)))),
\ea
$$
which is equivalent to first-order sentence 
\beq
\ba l
\forall x(p(x) \lrar x=a) \land \forall x(q(x) \lrar x=b)\\
\quad\land \forall x (r(x) \lrar (p(x) \land \neg q(x)))
\ea
\eeq{ex3f-comp} (see \cite{fer07a}, Example~3). The stable models of
$F$ are any first-order models of (\ref{ex3f}) whose signature is 
$\sigma(F)$. On the other hand $F$ has only one answer set:
$\{p(a),\ q(b),\ r(a)\}$.
\end{example}

We call a formula {\sl negative} if every occurrence of every predicate
constant in it belongs to the antecedent of an implication.
For instance, any formula of the form $\neg F$ is negative, because this
expression is shorthand for $F\rar\bot$.


\section{First-Order Loop Formulas}

\subsection{Review of Loop Formulas from~\cite{che06}}

We reformulate the definition of a first-order loop formula for
a nondisjunctive program from~\cite{che06}.

Let $\Pi$ be a nondisjunctive program that has no function constants
of positive arity, consisting of a finite number of
rules of the form \beq
   A\ar B, N
\eeq{abn-nd}
where $A$ is an atom, and $B$ is a set of atoms, and $N$ is a negative
formula.

We will say that $\Pi$ is in {\em normal form} if, for all rules
(\ref{abn-nd}) in~$\Pi$, no object constants occur in~$A$. It
is clear that every program can be turned into normal form using
equality. Let's assume that $\Pi$ is in normal form.

Let $\sigma(\Pi)$ be the signature consisting of object and predicate 
constants occurring in $\Pi$. Given a finite set~$Y$ of
non-equality atoms of $\sigma(\Pi)$, we first rename variables
in~$\Pi$ so that no variables occur in $Y$.
The {\em (first-order) external support formula} of $Y$ for $\Pi$,
denoted by $\fes_\Pi(Y)$, is the disjunction of
\beq
  \bigvee_{\theta : A\theta\in Y} \exists {\bf z}
     \bigg(B\theta\land N\theta\land
         \bigwedge_{p({\bf t})\in B\theta \atop p({\bf t}')\in Y} ({\bf t}\ne {\bf t}')
     \bigg)
\eeq{fes-nondis}
for all rules (\ref{abn-nd}) in~$\Pi$ where $\theta$ is a
substitution that maps variables in $A$ to terms occurring in $Y$,
and ${\bf z}$ is the list of all variables that occur in 
\[
   A\theta\ar B\theta, N\theta 
\]
but not in $Y$. \footnote{%
For any lists of terms ${\bf t}=(t_1,\dots,t_n)$
and ${\bf t}'=(t'_1,\dots,t'_n)$ of the same length, ${\bf t}={\bf t}'$ stands
for $t_1=t'_1\land \cdots \land t_n=t'_n$.}

The {\sl (first-order) loop formula} of $Y$, denoted by $\flf_\Pi(Y)$, is
the universal closure of
\beq
   \bigwedge Y \rar\fes_\Pi(Y) .
\eeq{y-fes-pi}

If $\Pi$ is a propositional program, for any nonempty finite set $Y$
of propositional atoms, $\flf_\Pi(Y)$ is equivalent to conjunctive
loop formulas defined in~\cite{fer06}, which we will denote by $\i{LF}_\Pi(Y)$.

The definition of a {\sl (first-order) loop} is as follows. We say that
$p({\bf t})$ {\sl depends on} $q({\bf t'})$ in~$\Pi$ if $\Pi$ has
a rule~(\ref{abn-nd}) such that
$p({\bf t})$ is $A$ and $q({\bf t'})$ is in $B$.
The {\em (first-order) dependency graph} of~$\Pi$ is an infinite
directed graph $(V,E)$ such that
\begin{itemize}
\item  $V$ is a set of non-equality atoms formed from $\sigma(\Pi)$,
       along with an infinite supply of variables;

\item  $(p({\bf t})\theta, q({\bf t'})\theta)$ is in $E$ if $p({\bf t})$
       depends on $q({\bf t'})$ in~$\Pi$ and $\theta$ is a substitution
       that maps variables in ${\bf t}$ and ${\bf t}'$ to object constants and
       variables occurring in~$V$.
\end{itemize}

A nonempty finite subset $L$ of $V$ is called a {\sl (first-order)
loop} of $\Pi$ if the subgraph of the first-order dependency graph of
$\Pi$ induced by $L$ is strongly connected.

\begin{example}\label{ex:1}
Let $\Pi$ be the following program:
\beq
\ba l
  p(x) \ar q(x) \\
  q(y) \ar p(y) \\
  p(x) \ar \no\ r(x).
\ea
\eeq{pqr}
The following sets are first-order loops: $Y_1=\{p(z)\}$,
$Y_2=\{q(z)\}$, $Y_3=\{r(z)\}$, $Y_4=\{p(z),q(z)\}$. Their loop
formulas are
\[
\ba {rcl}
\flf_{\Pi}(Y_1) &\!\!=\!\!& \forall z (p(z)\rar (q(z)\lor\neg r(z))); \\
\flf_{\Pi}(Y_2) &\!\!=\!\!& \forall z (q(z)\rar p(z)); \\
\flf_{\Pi}(Y_3) &\!\!=\!\!& \forall z (r(z)\rar\bot); \\
\flf_{\Pi}(Y_4) &\!\!=\!\!& \forall z (p(z)\land q(z)\rar\\
      & & (q(z)\land z\neq z)\lor (p(z)\land z \neq z)\lor \neg r(z)).
\ea
\]
\end{example}

\begin{example}\label{ex:2}
Let $\Pi$ be the one-rule program
\beq
   p(x)\ar p(y).
\eeq{p-xy}
Its first-order loops are $Y_k=\{p(x_1),\ldots,p(x_k)\}$ where $k>0$.
Formula $\flf_{\Pi}(Y_k)$ is
\beq
\ba l
   \forall x_1\ldots x_k (p(x_1)\land\ldots\land p(x_k)\\
   \qquad \rar\exists y (p(y)\land (y\ne x_1)\land\ldots\land (y\ne x_k))).
\ea
\eeq{flf-pi2}
\end{example}

\begin{definition} [Grounding a program] For any nondisjunctive program $\Pi$
we denote by $\i{Ground}(\Pi)$ the ground instance of $\Pi$, that is
the program obtained from~$\Pi$ by replacing every occurrence of
object variables with every object constant occurring in~$\Pi$, and
then replacing equality $a=b$ with $\top$ or $\bot$ depending on whether
$a$ is the same symbol as $b$.
\end{definition}

Given a program $\Pi$, let $(\sigma(\Pi))^g$ be a propositional
signature consisting of all the ground atoms of $\sigma(\Pi)$.
An Herbrand model of $\sigma(\Pi)$ can be identified with a
corresponding propositional model of $(\sigma(\Pi))^g$.

The following is a reformulation of Theorem~1 from~\cite{che06}.

\begin{prop} \label{prop:folf-nd}
Let $\Pi$ be a nondisjunctive program in normal form, and let $I$ be an
Herbrand model of~$\Pi$ whose signature is $\sigma(\Pi)$.
The following conditions are equivalent to each other:
\begin{itemize}
\item[(a)]  $I$ is an answer set of $\Pi$;

\item[(b)]  $I$ is an Herbrand model of \\
$\{\flf_\Pi(Y) :  Y \text{ is a nonempty finite set of atoms of }
\sigma(\Pi) \}$;

\item[(c)]  $I$ is an Herbrand model of \\
$\{\flf_\Pi(Y) : Y \text{ is a first-order loop of } \Pi \}$;

\item[(d)]  $I$ is a (propositional) model of
$\{\i{LF}_{\text{Ground}(\Pi)}(Y) :
Y \text{ is a nonempty (finite) set of ground atoms of } (\sigma(\Pi))^g \};
$

\item[(e)]  $I$ is a (propositional) model of
$\{\i{LF}_{\text{Ground}(\Pi)}(Y) : 
Y\text{ is a loop of }\i{Ground}(\Pi) \}\cup\ \{\neg p : p \text{ is an atom in } \\ (\sigma(\Pi))^g\text{ not occurring in } \i{Ground}(\Pi) \} .
$
\end{itemize}
\end{prop}

The sets of first-order loop formulas considered in conditions (b),
(c) above have obvious redundancy. For instance, the loop formula of
$\{p(x)\}$ is equivalent to the loop formula of $\{p(y)\}$;
the loop formula of $\{p(x),p(y)\}$ entails the loop formula of
$\{p(z)\}$. Following \cite{che06}, given two sets of atoms
$Y_1$, $Y_2$ not containing equality, we say that $Y_1$ {\em subsumes} $Y_2$ if
there is a substitution $\theta$ that maps variables in $Y_1$ to terms
so that $Y_1\theta=Y_2$. We say that $Y_1$ and $Y_2$ are {\em equivalent}
if they subsume each other.
\begin{prop}~\cite[Proposition~7]{che06}\label{prop:subsume}
Given two loops $Y_1$ and $Y_2$,
if $Y_1$ subsumes $Y_2$, then $\flf_\Pi(Y_1)$ entails $\flf_\Pi(Y_2)$.
\end{prop}

Therefore in condition (c) from Proposition~\ref{prop:folf-nd}, it is
sufficient to consider a set $\Gamma$ of loops such that for every
loop $L$ of $\Pi$, there is a loop $L'$
in $\Gamma$ that subsumes $L$. \citeauthor{che06}
[\citeyear{che06}] called this set of loops {\sl complete}. In
Example~\ref{ex:1}, set $\{Y_1,Y_2,Y_3,Y_4\}$ is a finite complete
set of loops of program~(\ref{pqr}). Program (\ref{p-xy}) in
Example~\ref{ex:2} has no finite complete set of loops.

In condition~(c) of Proposition~\ref{prop:folf-nd}, instead of
the first-order loops of the given program,
one may consider the first-order loops of any strongly equivalent
program, including a program that is not in normal form. This sometimes
yields a smaller number of loop formulas to consider. For example, 
the ground loops of program
\beq
\ba l
   p(a)\ar p(b) \\
   p(b)\ar p(c)
\ea
\eeq{p-abc}
are $\{p(a)\}$, $\{p(b)\}$, $\{p(c)\}$, all of which are subsumed by
$\{p(x)\}$. Thus it is sufficient to consider the loop formula of $\{p(x)\}$:
\beq
\ba l
   \forall x (p(x)\rar ((x=a)\land p(b)\land (x\ne b)) \\
\qquad\qquad\quad  \lor((x=b)\land p(c)\land (x\ne c))).
\ea
\eeq{p-abc-lf}
On the other hand, the ground loops of its normal form
\[
\ba l
  p(x)\ar x\!=\!a,\ p(b) \\
  p(x)\ar x\!=\!b,\ p(c)
\ea
\]
contain $\{p(b), p(c)\}$ in addition to the singleton ground loops.

\subsection{Extension to Disjunctive Programs}

A disjunctive program consists of a finite number of rules of the form
\beq
  A\ar B, N
\eeq{abn-d} where $A$, $B$ are sets of atoms, and $N$ is a
negative formula. As in the nondisjunctive case we assume that there
are no function constants of positive arity. Similar to above,
a program is in {\sl normal form} if, for all rules (\ref{abn-d})
in $\Pi$, no object constants occur in~$A$. We assume that $\Pi$
is in normal form.

Given a finite set $Y$ of non-equality atoms of~$\sigma(\Pi)$, 
we first rename variables in $\Pi$ so that no
variables occur in $Y$. The {\em (first-order) external support
formula} of $Y$ for $\Pi$, denoted by $\fes_\Pi(Y)$, is the
disjunction of
\begin{eqnarray}
 \bigvee_{\theta : A\theta\cap Y\ne\emptyset}\exists {\bf z}\hspace{-1.5em} &\hspace{-3em}
   \Big(B\theta\land N\theta\land\bigwedge_{p({\bf t})\in B\theta\atop p({\bf t}')\in Y} ({\bf t}\ne {\bf t}') \notag \\
   &\land\neg\big(\bigvee_{p({\bf t})\in A\theta} \big(p({\bf t})\land\bigwedge_{p({\bf t}')\in Y} {\bf t}\ne {\bf t}'\big)\big)\Big) \label{fes-dis}
\end{eqnarray}
for all rules (\ref{abn-d}) in~$\Pi$ where $\theta$ is a substitution that
maps variables in $A$ to terms occurring in $Y$ or to themselves, and
${\bf z}$ is the list of all variables that occur in
\[
   A\theta\ar B\theta, N\theta
\]
but not in $Y$. The {\sl (first-order) loop formula} of $Y$ for $\Pi$,
denoted by $\flf_\Pi(Y)$, is the universal closure
of~(\ref{y-fes-pi}). Clearly, (\ref{fes-dis}) is equivalent
to~(\ref{fes-nondis}) when $\Pi$ is nondisjunctive.


Similar to the nondisjunctive case, we say that $p({\bf t})$
{\sl depends on} $q({\bf t'})$ in
$\Pi$ if there is a rule~(\ref{abn-d}) in $\Pi$ such that
$p({\bf t})$ is in $A$ and $q({\bf t'})$ is in $B$. The notions of
grounding, a dependency graph and a first-order loop are extended to
disjunctive programs in a straightforward way.
Propositions~\ref{prop:folf-nd} and \ref{prop:subsume} can be extended to
disjunctive programs with these extended notions.

\begin{example} Let $\Pi$ be the following program
\hbox{$ p(x,y)\lor p(y,z) \ar q(x) $} and let $Y=\{p(u,v)\}$.
Formula $\flf_{\Pi}(Y)$ is the universal closure of
\[
\ba {ll}
p(u,v)\rar &
   \exists z(q(u)\land\neg (p(v,z) \land ((v,z)\ne (u,v)))) \\
   &\lor\ \exists x (q(x)\land\neg (p(x,u)\land ((x,u)\ne (u,v)))).
\ea
\]
\end{example}

\subsection{Extension to Arbitrary Sentences}

First-order loop formulas can even be extended to arbitrary sentences
under the stable model semantics~\cite{fer07a}.

As in~\cite{fer06}, it will be easier to discuss the
result with a formula whose {\sl negation} is similar to $\fes$.
We define formula $\fnes_F(Y)$ ({\sl ``Negation'' of $\fes$}) as follows,
where $F$ is a first-order formula and $Y$ is a finite set of atoms
not containing equality.
The reader familiar with~\cite{fer06} will notice that this is a
generalization of the notion $\i{NES}$ from that paper to
first-order formulas.

We assume that no variables in $Y$ occur in~$F$ by renaming bound
variables in $F$.
\begin{itemize}
\item  $\fnes_{p_i({\bf t})}(Y) =
    p_i({\bf t})\land\bigwedge_{p_i({\bf t'})\in Y} {\bf t}\ne {\bf t}'$;

\item  $\fnes_{t_1=t_2}(Y) = (t_1\!=\!t_2)$;
\item  $\fnes_{\bot}(Y) = \bot$;
\item  $\fnes_{F\land G}(Y) = \fnes_F(Y)\land\fnes_G(Y)$;
\item  $\fnes_{F\lor G}(Y) = \fnes_F(Y)\lor\fnes_G(Y)$;
\item  $\fnes_{F\rar G}(Y) = (\fnes_F(Y)\!\rar\!\fnes_G(Y))\land (F\!\!\rar\!\! G)$;
\item  $\fnes_{\forall x G}(Y) = \forall x \fnes_G(Y)$;
\item  $\fnes_{\exists x G}(Y) = \exists x \fnes_G(Y)$.
\end{itemize}

The (first-order) loop formula of $Y$ for sentence~$F$, denoted
by $\flf_F(Y)$, is the universal closure of
\beq
  \bigwedge Y\rar\neg\fnes_F(Y).
\eeq{folf-s}
It is not difficult to check that for any propositional formula~$F$
and any nonempty finite set $Y$ of propositional atoms, $\flf_F(Y)$ is
equivalent to $\i{LF}_F(Y)$, where $\i{LF}$ denotes
loop formula for a propositional formula as defined in~\cite{fer06}.

This notion of a loop formula is a generalization of a loop formula
for a disjunctive program in view of the following lemma.

\begin{lemma}
Let $\Pi$ be a disjunctive program in normal form, $F$ the
FOL-representation of $\Pi$, and $Y$ a finite set of atoms not
containing equality. Formula $\fnes_F(Y)$ is equivalent to
$\neg\fes_\Pi(Y)$ under the assumption~$\Pi$.
\end{lemma}

To define a first-order dependency graph of $F$, we need a few notions.
Recall that an occurrence of a formula $G$ in a formula~$F$ is {\em
  positive} if the number of implications in~$F$ containing that
occurrence in the antecedent is even; it is {\em strictly positive}
if that number is~$0$.
%
We will call a formula in {\em rectified form} if it has no variables
that are both bound and free, and the quantifiers are followed by
pairwise distinct variables. Any formula can be turned into rectified
form by renaming bound variables.

Let $F$ be a formula in rectified form.
We say that an atom $p({\bf t})$ {\em weakly depends on} an atom $q({\bf t'})$
in an implication $G\rar H$ if
\begin{itemize}
\item  $p({\bf t})$ has a strictly positive occurrence in~$H$, and
\item  $q({\bf t'})$ has a positive occurrence in~$G$ that does not
       belong to any occurrence of a negative formula in~$G$.
\end{itemize}

We say that $p({\bf t})$ {\em depends on} $q({\bf t'})$ in~$F$ if
$p({\bf t})$ weakly depends on $q({\bf t'})$ in an implication that
has a strictly positive occurrence in~$F$.

The definition of a first-order dependency graph for a nondisjunctive
program is extended to $F$ in a straightforward way using this
extended notion of dependency between two atoms.
A loop is also defined similarly.

\begin{definition}[Grounding a sentence]
For any sentence $F$ that has no function constants of positive arity,
$\i{Ground}(F)$ is defined recursively. If $F$ is an atom
$p({\bf t})$ then $\i{Ground}(F)$ is $F$.  If $F$ is an equality $a=b$
then $\i{Ground}(F)$ is $\top$ or $\bot$ depending on whether $a$ is
the same symbol as $b$. The function $\i{Ground}$ commutes with all
propositional connectives; quantifiers turn into finite conjunctions
and disjunctions over all object constants occurring in $F$.
\end{definition}

Proposition~\ref{prop:folf-nd} remains correct even after replacing
``a nondisjunctive program in normal form'' in the statement with
``a sentence in rectified form that contains no function constants of
positive arity,'' and using the extended notions accordingly.
Proposition \ref{prop:subsume} can be extended to arbitrary sentences as well.


\section{Loop Formulas in Second-Order Logic}

\subsection{$\sm$ and Loop Formulas}

Let $F$ be a first-order formula, let $p_1,\dots, p_n$ be the
list of all predicate constants occurring in $F$, and let ${\bf u}$ and
${\bf v}$ be lists of predicate variables corresponding to
$p_1,\dots,p_n$.

We define $\snes_F({\bf u})$ recursively as follows, which is similar
to $\fnes$ above but contains second-order variables as its argument.
%
\begin{itemize}
\item  $\snes_{p_i({\bf t})}({\bf u}) =
       p_i({\bf t}) \land \neg u_i({\bf t})$;

\item  $\snes_{t_1=t_2}({\bf u}) = (t_1\!=\!t_2)$;

\item  $\snes_{\bot}({\bf u}) = \bot$;

\item  $\snes_{F\land G}({\bf u}) = \snes_F({\bf u})\land\snes_G({\bf u})$;

\item  $\snes_{F\lor G}({\bf u}) = \snes_F({\bf u})\lor\snes_G({\bf u})$;

\item  $\snes_{F\rar G}({\bf u}) =
          (\snes_F({\bf u})\!\rar\!\snes_G({\bf u}))\land (F\!\!\rar\! G)$;

\item  $\snes_{\forall x F}({\bf u}) = \forall x \snes_F({\bf u})$;

\item  $\snes_{\exists x F}({\bf u}) = \exists x \snes_F({\bf u})$.
\end{itemize}

By $\NE({\bf u})$ we denote the formula
\[
   \exists {\bf x}^1 u_1({\bf x}^1)\lor\dots\lor
   \exists {\bf x}^n u_n({\bf x}^n).
\]

$\sm[F]$ can be written in the style of ``loop formulas'' in the
following way.
\begin{prop}\label{prop:2nd-order}
For any sentence $F$, $\sm[F]$ is equivalent to
\beq
   F\land\forall {\bf u}(({\bf u}\le {\bf p})
                 \land\NE({\bf u})\rar\neg\snes_F({\bf u})) .
\eeq{lf2-ne}
\end{prop}

\subsection{Second-Order Characterization of Loops}

The notion of a loop can be incorporated into the second-order definition
of stable models as follows.

Given a sentence $F$ in rectified form, by $E_F({\bf v},{\bf u})$ we denote
\[
   \bigvee_{(p_i({\bf t}),p_j({\bf t'}))\; :\; \atop
     p_i({\bf t}) \text{ depends on } p_j({\bf t'}) \text{ in } F}
     \exists {\bf z} (v_i({\bf t})\land u_j({\bf t'})\land\neg v_j({\bf t'}))
\]
where ${\bf z}$ is the list of all object variables in ${\bf t}$ and
${\bf t'}$.
By $\i{SC}_F({\bf u})$ we denote the second-order sentence
\beq
   \NE({\bf u})\land\forall {\bf v} (({\bf v}<{\bf u})\land\NE({\bf v})
            \rar E_F({\bf v},{\bf u})).
\eeq{scf}
Formula~(\ref{scf}) represents the concept of a loop without referring to
the notion of a dependency graph explicitly, based on the following 
observation. Consider a finite propositional program $\Pi$. 
A set $U$ of atoms is a loop of $\Pi$ iff for every nonempty proper 
subset $V$ of $U$, there is an edge from an atom
in $V$ to an atom in $U\setminus V$ in the dependency graph of
$\Pi$ \cite{geb06}.
To see the relation in the first-order case, we first define a dependency
graph and a loop that are relative to a given interpretation.
Let $F$ be a sentence in rectified form and let $I$ be an interpretation
of $F$. The {\em dependency graph of $F$ w.r.t.~$I$} is an infinite
directed graph $(V,E)$ where
\begin{itemize}
\item  $V$ is the set of all atoms of the form $p_i(\vec{\xi^*})$ where
       $\vec{\xi^*}$ is a list of object names,~\footnote{Each element
       $\xi$ of the universe $|I|$ has a corresponding {\sl object name}, 
       which is an object constant not from the given signature
       See~\cite{lif08} for details.}
       and

\item  $(p_i(\vec{\xi^*}), p_j(\vec{\eta^*}))$ is in $E$
   if there are $p_i({\bf t})$, $p_j({\bf t}')$ such that
   $p_i({\bf t})$ depends on $p_j({\bf t}')$ in $F$ and
   there is a mapping $\theta$ from variables in ${\bf t}$ and ${\bf t}'$
   to object names such that $({\bf t}\theta)^I = \vec{\xi}$, and
   $({\bf t}'\theta)^I = \vec{\eta}$.
\end{itemize}

We call a nonempty subset $L$ of $V$ a {\sl loop} of $F$ w.r.t.~$I$ if the
subgraph of the dependency graph of $F$ w.r.t.~$I$ that is induced by $L$
is strongly connected.\footnote{Note that unlike first-order loops defined
earlier we don't restrict $L$ to be finite. There the assumption was 
required to be able to write a loop formula.}
The following lemma describes the relation between formula (\ref{scf}) 
and a loop w.r.t.~$I$.

\begin{lemma}\label{lem:scf-sc}
Let $F$ be a first-order sentence in rectified form, $I$ an interpretation
of $F$ and ${\bf q}$ a list of predicate names \footnote{%
Like object names, for every $n>0$, each subset of $|I|^n$ has a name, 
which is an $n$-ary predicate constant not from the given signature.}
corresponding to ${\bf p}$.
$I\models\i{SC}_F({\bf q})$ iff
\[
  Y=\{p_i(\vec{\xi^*}) : q_i^I(\vec{\xi})\!\!=\!\!\text{\sc true} 
     \text{ where }\vec{\xi}\text{ is a list of object names}\}
\]
is a loop of $F$ w.r.t.~$I$.
\end{lemma}

One may expect that, similar to the equivalence between conditions
(b) and (c) from Proposition~\ref{prop:folf-nd}, formula
(\ref{lf2-ne}) is equivalent to the following formula: \beq
   F\land\forall {\bf u}(({\bf u}\le {\bf p})
                 \land\i{SC}_F({\bf u})\rar\neg\snes_F({\bf u})) .
\eeq{lf2-sc}
However, this is not the case as shown in the following example.

\begin{example}\label{ex:unbounded}
Let $F$ be the FOL-representation of program~$\Pi:$
\[
\ba l
  p(x,y) \ar q(x,z) \\
  q(x,z) \ar p(y,z).
\ea
\]
Consider interpretation $I$ whose universe is the set of all nonnegative 
integers such that
\[
\ba l
   p^I(m,n)= \begin{cases}
               \text{{\sc true}} & \text{ if } m=n, \\
               \text{{\sc false}} & \text{ otherwise; }
             \end{cases} \\
   q^I(m,n) = \begin{cases}
               \text{{\sc true}} & \text{ if } n=m+1, \\
               \text{{\sc false}} & \text{ otherwise; }
             \end{cases}
\ea
\]
One can check that $I$ is not a stable model of $F$, but
satisfies~(\ref{lf2-sc}).

\end{example}

This mismatch is similar to the observation from~\cite{lee05} that
the external support of all loops does not ensure the stability of the
model if the program is allowed to be infinite. Consider the following
infinite program:
\beq
  p_i \ar p_{i+1} \qquad (i>0).
\eeq{ex:infinite}
The only loops are singletons, and their loop formulas are satisfied
by the model $\{p_1,p_2,\ldots\}$ of (\ref{ex:infinite}), which is not
stable. To check the stability, not only we need to check every loop 
is externally supported, but also need to check that
$\{p_1,p_2,\ldots\}$ is ``externally supported.''
Example~\ref{ex:unbounded} shows that the mismatch can occur even if
the program is finite once it is allowed to contain variables.
What distinguishes $\{p_1,p_2,\ldots\}$ from loops is that, for 
every loop in $\{p_1,p_2,\ldots\}$, there is an outgoing edge in the
dependency graph. Taking this into account, we define 
$\i{Loop}_F({\bf u})$ as
\beq
\ba l
   \i{SC}_F({\bf u}) \lor (\NE({\bf u}) \\
\qquad\qquad\qquad    \land\ \forall {\bf v}  
    (({\bf v}\le {\bf u})\land\i{SC}_F({\bf v})
            \rar E_F({\bf v},{\bf u}))).
\ea
\eeq{loopf}

Given a dependency graph of $F$ w.r.t.~$I$, we say that a nonempty set $Y$ 
of vertices is {\em unbounded} w.r.t.~$I$ if, for every subset $Z$ of
$Y$ that induces a strongly connected subgraph, there is an edge from
a vertex in $Z$ to a vertex in $Y\setminus Z$. For instance, for the
interpretation $I$ in Example~\ref{ex:unbounded}, 
\[ 
  \{ p(0^*,0^*), q(0^*, 1^*), p(1^*, 1^*), q(1^*, 2^*),\ldots,\}
\]
is an unbounded set w.r.t.~$I$.

The following lemma describes the relation between the
second disjunctive term of (\ref{loopf}) with unbounded sets.

\begin{lemma}\label{lem:unboundedf-unbounded}
Let $F$ be a first-order sentence in rectified form, $I$ an interpretation,
and ${\bf q}$ a list of predicate names corresponding to ${\bf p}$.
\[ I\models \NE({\bf q})\land
      \forall {\bf v} (({\bf v}\le {\bf q})\land\i{SC}_F({\bf v})
            \rar E_F({\bf v},{\bf q}))
\]
iff 
\[
  Y=\{p_i(\vec{\xi^*}) : q_i^I(\vec{\xi})\!\!=\!\!\text{\sc true} 
     \text{ where }\vec{\xi}\text{ is a list of object names}\}
\]
is an unbounded set of $F$ w.r.t.~$I$.
\end{lemma}

An {\em extended loop} of $F$ w.r.t.~$I$ is a loop or an unbounded set
of $F$ w.r.t.~$I$.
Clearly $I\models (\ref{loopf})$ iff  
\[
  Y=\{p_i(\vec{\xi^*}) : q_i^I(\vec{\xi})\!\!=\!\!\text{\sc true} 
     \text{ where }\vec{\xi}\text{ is a list of object names}\}
\]
is an extended loop of $F$ w.r.t.~$I$.

The following proposition shows that the formula obtained from
(\ref{lf2-sc}) by replacing $\i{SC}_F({\bf u})$ with
$\i{Loop}_F({\bf u})$ is equivalent to $\sm[F]$.

\medskip\noindent{\sl
{\bf Proposition~\ref{prop:2nd-order}$'$}\ \ For any sentence $F$ in
rectified form, the following second-order sentences are equivalent to
each other:
\begin{itemize}
\item[$(a)$]  $\sm[F]$;
\item[$(b)$]  $F\land\forall {\bf u}(({\bf u}\le {\bf p})\land\NE({\bf u})
                    \rar\neg\snes_F({\bf u}))$;
\item[$(c)$]  $F\land\forall {\bf u}(({\bf u}\le {\bf p})\land\loop_F({\bf u})
                    \rar\neg\snes_F({\bf u}))$.
\end{itemize}
}
\noindent
(See appendix A for an example.)

\medskip
Proposition~\ref{prop:2nd-order}$'$ is essentially a generalization of
the main theorem from~\cite{fer06} to first-order sentences.
If $F$ is a propositional formula, then for any subset $Y$ of {\bf p},
by $\oa Y$ we denote the tuple $(Y_1,\dots,Y_n)$, where
$$
Y_i =
\left\{
 \ba{ll}
   \top, & \hbox{if $p_i\in Y$}; \\
   \bot, & \hbox{otherwise}.
 \ea
\right.
$$

\begin{cor}\cite[Theorem~2]{fer06}
For any propositional formula $F$, the following conditions are equivalent
to each other under the assumption $F$.
\begin{itemize}
\item[(a)]  $\sm[F]$;

\item[(b)]  The conjunction of $\bigwedge Y\rar\neg\snes_F(\oa{Y})$
            for all nonempty sets $Y$ of atoms occurring in $F$;

\item[(c)]  The conjunction of $\bigwedge Y\rar\neg\snes_F(\oa{Y})$
            for all loops $Y$ of $F$.
\end{itemize}
\end{cor}

Several other propositions in this paper are derived from
Proposition~\ref{prop:2nd-order}$'$.


\section{Between $\sm$ and First-Order Loop Formulas}

In general, $\sm[F]$ is not reducible to any first-order sentence, even
in the absence of function constants of positive arity. As
in circumscription, transitive closure can be represented using $\sm$,
while it cannot be done by any set of first-order formulas, even if that set
is allowed to be infinite.\footnote{Vladimir Lifschitz, personal communication.}
However, if the universe consists of finite elements, then the
following holds.
We will say that $F$ is in {\sl normal form} if no object constants
occur in a strictly positive occurrence of atoms in $F$.

\begin{prop}\label{prop:1st-2nd}
For any sentence $F$ and any model $I$ of $F$ whose universe is
finite, the following conditions are equivalent:
\begin{itemize}
\item[(a)]  $I$ satisfies $\sm[F]$;


\item[(b)]  for every nonempty finite set $Y$ of atoms formed
            from predicate constants in $\sigma(F)$ and an infinite
            supply of variables, $I$ satisfies $\flf_F(Y)$.
\end{itemize}
If $F$ is in rectified and normal form that has no function constants
of positive arity, the following condition is also equivalent to each
of (a) and (b):
\begin{itemize}
\item[(c)]  for every first-order loop $Y$ of $F$,
            $I$ satisfies $\flf_F(Y)$.
\end{itemize}
\end{prop}

Unlike Proposition~\ref{prop:folf-nd} in which loops can be found from
any strongly equivalent program, condition (c) requires that loops
be found from a normal form. This is related to the fact that 
Proposition~\ref{prop:1st-2nd} considers non-Herbrand stable models 
as well, which may not satisfy the unique name assumption. For instance, 
recall that program (\ref{p-abc}) has singleton loops only, which are 
subsumed by $\{p(x)\}$.
Consider an interpretation $I$ such that $|I|=\{1,2\}$ and 
$a^I = c^I = 1$, $b^I = 2$, $p^I(m)=\text{\sc true}$ for $m=1,2$.
$I$ is a non-Herbrand model which is not stable, but it satisfies 
(\ref{p-abc-lf}), the loop formula of $\{p(x)\}$.

The proof of the equivalence between (a) and (c) uses the following lemma.

\begin{lemma}\label{lem:no-inf-extloop}
Let $F$ be a sentence in rectified and normal form that contains 
no function constants of positive arity, and let $I$ be an interpretation.
If there is no infinite extended loop of $F$
w.r.t. $I$, then $I \models \sm[F]$ iff, for every first-order loop
$Y$ of $F$, $I\models\flf_F(Y)$.
\end{lemma}

Without the finite universe assumption, Proposition~\ref{prop:1st-2nd}
would be incorrect, as shown in Example~\ref{ex:unbounded}. For
another example, consider program~(\ref{p-xy}) with an
interpretation $I$ with an infinite universe such that $p$ is identically true.
$I$ does not satisfy $\sm[F]$, but satisfies $F$ and $\flf_F(Y)$ for
any finite set $Y$ of atoms.

In view of Proposition~\ref{prop:subsume}, if the size of the universe
$|I|$ is known, as with the answer sets (whose universe is the
Herbrand universe of~$\sigma(F)$), it is sufficient to consider at
most $2^{|{\bf p}|}-1$ loop
formulas where ${\bf p}$ is set of all predicate constants occurring
in the sentence. Each loop formula is for set $Y_{\bf q}$ corresponding to
a nonempty subset ${\bf q}$ of ${\bf p}$, defined as
\hbox{$Y_{\bf q}=\{p({\bf x_1}),\ldots, p({\bf x_{|I|^n}})\ :\ p\in {\bf q}\}$}
where $n$ is the arity of~$p$. 
For instance, for program~(\ref{p-xy}), if the size
of the universe is known to be $3$, it is sufficient to consider only one
loop formula (\ref{flf-pi2}) where $k=3$.

In the next section we consider certain classes of sentences for which
$\sm[F]$ is equivalent to a first-order sentence without the finite
universe assumption.


\section{Reducibility to first-order formulas}\label{sec:reduce}

\subsection{Finite complete set of first-order loops}

Proposition~8 from~\cite{fer07a} shows that $\sm[F]$ can be reduced
to a first-order sentence if $F$ is ``tight'', i.e., $F$ has no
``nontrivial'' predicate loops. (Predicate loops are defined similar
to first-order loops, but from a ``predicate dependency
graph''~\cite{fer07a}, which does not take into account ``pointwise
dependency.'')  We further generalize this result using the notion of
finite complete set of loops.

Let $F$ be a sentence in rectified form that contains no function
constants of positive arity. Theorem~2 from~\cite{che06} provides a
syntactic condition under which a nondisjunctive program has a finite
complete set of loops, which can be extended to disjunctive programs
and arbitrary sentences in a straightforward way.

The following proposition tells that if $F$ has a finite complete set
of loops, then $\sm[F]$ can be equivalently rewritten as a first-order
sentence.

\begin{prop}\label{prop:fcsl}
Let $F$ be a sentence in rectified and normal form that contains
no function constants of positive arity. 
If $F$ has a finite complete set $\Gamma$ of first-order loops,
then $\sm[F]$ is equivalent to the conjunction of $F$ with the set of loop
formulas for all loops in $\Gamma$.
\end{prop}

This proposition generalizes Proposition~8 from~\cite{fer07a}.
Clearly, every tight sentence has a finite complete set of first-order loops.


The proof of Proposition~\ref{prop:fcsl} follows from
Lemma~\ref{lem:no-inf-extloop} and the following lemma.

\begin{lemma}\label{lem:fcs-relative}
Let $F$ be a sentence in rectified and normal form that contains
no function constants of positive arity. 
If $F$ has a finite complete set of loops, then there is no infinite
extended loop of $F$ w.r.t any interpretation.
\end{lemma}

Proposition~\ref{prop:fcsl} would go wrong if we replace ``a finite
complete set of loops'' in the statement with ``a finite number of
predicate loops.'' Obviously any sentence~$F$ contains a finite number
of predicate constants, so that this condition is trivial. In view of
intranslatability of $\sm$ to first-order sentences, this fact tells
that the more refined notion of first-order loops is essential for
this proposition to hold.

For nondisjunctive program $\Pi$, Proposition~9 from~\cite{che06} shows
that if every variable in the head occurs in the body, then $\Pi$ has
a finite complete set of loops. However, this does not hold once $\Pi$
is allowed to be disjunctive. For instance,
\[
\ba l
   p(x,y)\ar q(x), r(y)  \\
   q(x)\lor r(y) \ar p(x,y)
\ea
\]
has no finite complete set of loops.

\subsection{Safe formulas}

A disjunctive program~$\Pi$ is called {\sl safe} if, for each
rule~(\ref{abn-d}) of $\Pi$, every variable occurring in the rule
occurs in $B$ as well. \cite{lee08} generalized this notion to
sentences, showing that for any safe sentence, its Herbrand stable
models are not affected by ``irrelevant'' object constants that do not
occur in the program. We will show that this notion is also related to
reducing $\sm[F]$ to a first-order sentence.

We review the notion of safety from~\cite{lee08}.\footnote{The
definition here is slightly weaker and applies to arbitrary sentences,
unlike the one in~\cite{lee08} that refers to prenex form.}
We assume that there are no function constants of positive arity.
As a preliminary step, we assign to every formula~$F$ in rectified
form a set $\i{RV}(F)$ of its {\sl restricted variables}, as follows:
\begin{itemize}
\item  For an atom~$F$,
  \begin{itemize}
  \item if~$F$ is an equality between two variables then
   $\i{RV}(F)=\emptyset$;
   \item otherwise, $\i{RV}(F)$ is the set of all variables
       occurring in $F$;
  \end{itemize}
\item  $\i{RV}(\bot)=\emptyset$;
\item  $\i{RV}(F\land G)=\i{RV}(F)\cup\i{RV}(G)$;
\item  $\i{RV}(F\lor G)=\i{RV}(F)\cap\i{RV}(G)$;
\item  $\i{RV}(F\rar G)=\emptyset$;
\item  $\i{RV}(Qv F)=\i{RV}(F)\setminus\{v\}$ where $Q\in\{\forall, \exists\}$.
\end{itemize}

We say that a variable $x$ is {\em unsafe} in $F$ if there is an
occurrence of $x$ in $F$ that is not in any of
\begin{itemize}
\item  $\forall x$, $\exists x$, and
\item  any subformula $G\rar H$ of $F$ such that $x\in\i{RV}(G)$.
\end{itemize}

By $U_F$ we denote the formula
\[
  \bigwedge_{p\in{\bf p}}\forall {\bf x} \Bigl(p({\bf x})\rar
        \bigwedge_{x\in {\bf x}} \bigvee_{c\in C} x=c \Bigr)
\]
where $C$ is the set of all object constants occurring in~$F$, and
${\bf x}$ is a list of distinct object variables whose length is
the same as the arity of $p$.

The following proposition tells that for a safe sentence $F$, formula
$\sm[F]$ can be equivalently rewritten as a first-order sentence.

\begin{prop}\label{prop:safety}
Let $F$ be a sentence in rectified form that has no function constants
of positive arity. If $F$ has no unsafe variables,
  then $\sm[F]$ is equivalent to the conjunction of $F$, $U_F$ and a finite
  number of loop formulas.
\end{prop}

We note that the syntactic conditions in Propositions~\ref{prop:fcsl}
and \ref{prop:safety} do not entail each other.
For instance, 
\hbox{$\forall x\ (q(x)\land p(y)\rar p(x))$}
\ \
has no unsafe variables, but has no finite complete set of first-order
loops, while 
$
  \forall x\ p(x)
$
\ \
has a finite complete set of loops  $\{\{p(x)\}\}$, but has an unsafe
variable~$x$.

Safety is usually imposed on input programs for
answer set solvers, but it could be somewhat restricted in first-order
reasoning which is not confined to generating Herbrand stable
models. For instance, the example program in the introduction
(identified as a sentence)
has an unsafe variable $w$ (but has a finite complete set of loops).


\section{Programs with Explicit Quantifiers}

In the following we extend the syntax of logic programs by allowing explicit
quantifiers. As in answer set programs, the syntax uses the 
intuitive if-then form, but allows explicit quantifiers.
An {\sl extended rule} is of the form
\beq
 H\ar G
\eeq{ex-rule}
where $G$ and $H$ are formulas with no function constants of positive
arity such that every occurrence of an implication in~$G$ and $H$ is
in a negative formula. 
An extended program is a finite set of extended rules.
The semantics of an extended program is defined by identifying the
program with $\sm[F]$ where $F$ is a conjunction of the universal
closure of implications that correspond to the rules (FOL-representation).
An example of an extended program is given in the introduction.

Let $\Pi$ be an extended program. 
Given a nonempty finite set $Y$ of non-equality atoms of
$\sigma(\Pi)$, we first rename variables in $\Pi$ so that no variables
occur in $Y$. 
Formula $\i{EFES}_\Pi(Y)$ ({\sl ``Extended $\fes$''}) is defined as
the disjunction of
\beq
   \exists {\bf z} (\fnes_G(Y)\land\neg\fnes_H(Y))
\eeq{efes}
for all rules (\ref{ex-rule}) where $H$ contains a strictly positive
occurrence of a predicate constant that belongs to $Y$, and ${\bf z}$
is the list of all free variables in the rule that do not occur in $Y$.

The loop formula of $Y$ for $\Pi$ is the universal closure of
\beq
   \bigwedge Y\rar\i{EFES}_\Pi(Y).
\eeq{lf-extended}

The following proposition tells that (\ref{lf-extended}) is a
generalization of the definition of a loop formula for a disjunctive
program and is equivalent to the definition of a loop formula
(\ref{folf-s}) for an arbitrary sentence. 

\begin{prop}\label{prop:lf-extended-program}  Let $\Pi$ be an extended
  program, $F$ the FOL-representation of $\Pi$, and $Y$ a finite set
  of atoms not containing equality. Under the assumption $\Pi$,
  formula~$\i{EFES}_\Pi(Y)$ is equivalent to $\neg\fnes_F(Y)$.
  If $\Pi$ is a disjunctive program, then $\i{EFES}_\Pi(Y)$ is also
  equivalent to $\fes_\Pi(Y)$ under the assumption $\Pi$.
\end{prop}

While the size of
(\ref{folf-s}) is exponential to the size of $F$ in the worst case,
(\ref{lf-extended}) can be equivalently written in a linear size due to the
following lemma.
\begin{lemma}\label{propneg}
For any negative formula $F$ and any finite set $Y$ of non-equality atoms, 
$\fnes_F(Y)$ is equivalent to $F$. 
\end{lemma}

For instance, for $F=(p(x)\!\!\rar\!\!\bot)\!\!\rar\!\!\bot$ and $Y=\{p(a)\}$, 
formula $\fnes_F(Y)$ is 
\[ 
  [(((p(x)\land x\!\ne\! a)\!\!\rar\!\!\bot)\land
     (p(x)\!\!\rar\!\!\bot))
          \!\!\rar\!\!\bot]
  \land [(p(x)\!\!\rar\!\!\bot)\!\!\rar\!\!\bot],
\]
which is equivalent to $F$. 

A finite set $\Gamma$ of sentences {\sl entails} a sentence $F$ under
the stable model semantics (symbolically, $\Gamma\smmodels F$), if
every stable model of $\Gamma$ satisfies $F$.

If $\sm[F]$ can be reduced to a first-order sentence, as allowed in
Propositions~\ref{prop:fcsl} and \ref{prop:safety}, clearly, the
following holds.
\[
   \Gamma\smmodels F \text{ iff } \Gamma\cup\Delta\models F
\]
where $\Delta$ is the set of first-order loop formulas required.
This fact allows us to use first-order theorem provers to reason about
query entailment under the stable model semantics.

\begin{example}\label{ex:intro}
Consider the insurance policy example in the introduction, which has
the following finite complete set of loops: 
$\{\i{Divorced}(u,v)\}$,
$\{\i{Accident}(u,v)\}$,
$\{\i{Discount}(u,v)\}$,
$\{\i{GotMarried}(u,v)\}$, $\{\i{Spouse}(u,v)\}$ and
$\{\i{GotMarried}(u,v), \i{Spouse}(u,v)\}$. 
Their loop formulas for $\Pi_1\cup\Pi_2$ are equivalent to the
universal closure of
\[
\ba l
Div(u,v)\rar\bot \\ \\

Acc(u,v)\rar\bot \\ \\

Dis(u,v)\rar 
   \exists xy [Spo(x,y)\land\neg\exists z Acc(x,z) \\
\qquad\qquad\qquad\land\ \neg(\exists w (Dis(x,w)\land (x,w)\ne
   (u,v)))] \\ \\
Mar(u,v)\rar \\
\qquad \exists xy [Spo(x,y)\land\neg(Mar(x,y)\land (x,y)\ne (u,v))] \\
\quad \lor\ \neg\exists  y [Mar(marge,y)\land (marge,y)\ne (u,v)]  \\ \\
Spo(u,v)\rar \\
\qquad \exists xy [Mar(x,y)\land\neg Div(x,y) \\ 
\qquad\qquad\qquad \land\neg (Spo(x,y)\land (x,y)\ne (u,v))] \\ \\

Mar(u,v)\land Spo(u,v)\rar \\
\qquad \exists xy [(Spo(x,y)\land (x,y)\ne (u,v)) \\
\qquad\qquad\qquad\land\neg(Mar(x,y)\land (x,y)\ne (u,v))] \\
\quad \lor\ \neg\exists  y [Mar(marge,y)\land (marge,y)\ne (u,v)]  \\ 
\quad \lor\ \exists xy [(Mar(x,y)\land (x,y)\ne (u,v))\land\neg Div(x,y) \\ 
\qquad\qquad\qquad \land\neg (Spo(x,y)\land (x,y)\ne (u,v))] .
\ea
\]
These loop formulas, conjoined with the FOL-representation of
$\Pi_1\cup\Pi_2$, entail under first-order logic each of $\exists xw\
Dis(x,w)$ and \hbox{$\forall x (Dis(x,\i{plan1})\rar x=\i{marge})$}.
We verified the answers using a first-order theorem
prover Vampire~\footnote{{\tt http://www.vampire.fm .}}.
\end{example}


\section{Conclusion}

Our main contributions are as follows.

\begin{itemize}

\item  We extended loop formulas with variables from~\cite{che06} to
  disjunctive programs and to arbitrary first-order sentences and showed
  their relations to the new language of stable models
  from~\cite{fer07a}.

\item  We presented certain syntactic conditions under which the
  language of stable models from~\cite{fer07a} can be reduced to
  first-order logic, which allows to use first-order theorem provers
  to reason about stable models.

\item  We defined the notion of an extended program which allows
  closed-world reasoning under the stable model semantics even in the
  absence of the unique name and the domain closure
  assumptions. 
  We provided a computational method for extended programs by means
  of loop formulas.
\end{itemize}

The use of first-order theorem provers for the stable model
semantics was already investigated in~\cite{sab07}, but their
results are limited in several ways. They considered nondisjunctive logic
programs with ``trivial'' loops only, in which case the stable model
semantics is equivalent to the completion semantics~\cite{cla78}; their
notion of models were limited to Herbrand models.

SAT-based answer set solvers may also benefit from loop formulas with
variables. Instead of finding propositional loop formulas one by one from
the ground program, one may consider a set of formulas in a batch which
are obtained from grounding first-order loop formulas. Whether it leads
to computational efficiency needs empirical evaluation.

\section*{Acknowledgements}

We are grateful to Vladimir Lifschitz, Ravi Palla and the anonymous
referees for their useful comments on this paper. 


\bibliographystyle{named}

\appendix

\section{Appendix. Additional Examples}

Consider program~(\ref{pqr}) from Example~\ref{ex:1}:
\[
\ba l
 p(x) \ar q(x) \\
 q(y) \ar p(y) \\
 p(x) \ar \no\ r(x).
\ea
\]
Let $F$ be the FOL-representation of $\Pi$:
\[
 \forall xy((q(x)\rar p(x))\land (p(y)\rar q(y))\land (\neg r(x)\rar p(x))).
\]

Below we use the following fact to simplify the formulas.
\begin{lemma}\label{propneg2}
For any negative formula $F$, the formula
$$\snes_F({\bf u})\lrar F$$
is logically valid.
\end{lemma}

\bigskip\noindent
{\bf 1. $\sm[F]$} is equivalent to
\[
\ba l
 F\land \neg \exists u_1 u_2 u_3 ((u_1,u_2,u_3)<(p,q,r)) \\
\qquad\qquad \land \forall xy ( (u_2(x)\rar u_1(x)) \\
\qquad\qquad\qquad\ \ \land     (u_1(y)\rar u_2(y)) \\
\qquad\qquad\qquad\ \ \land     (\neg r(x)\rar u_1(x)) ). \ea
\]

\bigskip\noindent
{\bf 2. Formula in Proposition~\ref{prop:2nd-order}$'$ (b):}
\[
 F\land\forall {\bf u}({\bf u}\le {\bf p}\land\NE({\bf u})
                   \rar\neg\snes_F({\bf u}))
\]
is equivalent to
\beq
\ba l
 F\land\forall u_1 u_2 u_3 ((u_1,u_2,u_3)\le (p,q,r) \\
\qquad\quad \land (\exists x\ u_1(x)\lor\exists x\ u_2(x)\lor\exists
x\ u_3(x))\\
\qquad\quad\quad\rar
  \neg\forall xy([q(x)\land\neg u_2(x)\rar p(x)\land\neg u_1(x)] \\
\qquad\qquad\qquad\quad \land
                [p(y)\land \neg u_1(y) \rar q(y)\land\neg u_2(y)] \\
\qquad\qquad\qquad\quad \land
                [\neg r(x)\rar p(x)\land\neg u_1(x)])).
\ea
\eeq{for-b}

\bigskip\noindent
{\bf 3. Formula in Proposition~\ref{prop:2nd-order}$'$ (c):} Similar to
(\ref{for-b}) except that
\[
  \exists x\ u_1(x)\lor\exists x\ u_2(x)\lor\exists x\ u_3(x)
\]
in (\ref{for-b}) is replaced with $\i{Loop}_F({\bf u})$, which is
\[
\ba l \i{SC}_F({\bf u}) \lor [(\exists x\ u_1(x)\lor \exists x\
u_2(x)\lor \exists x\ u_3(x)) \\
\land\ \forall v_1 v_2 v_3 ((
(v_1,v_2,v_3) \leq (u_1,u_2,u_3)) \land \i{SC}_F({\bf v}) \\
\qquad\qquad\quad\ \rar (\exists x(v_1(x)\land u_2(x)\land\neg v_2(x))\\
\qquad\qquad\qquad\quad\lor \exists y (v_2(y)\land u_1(y)\land\neg
v_1(y))))], \ea
\]

where $\i{SC}_F({\bf u})$ is 
\[
\ba l
(\exists x\ u_1(x)\lor \exists x\ u_2(x)\lor \exists x\ u_3(x)) \\
\land\ \forall v_1 v_2 v_3 (((\exists x\ v_1(x)\lor\exists x\
v_2(x)\lor
\exists x\ v_3(x)) \\
\qquad\qquad\quad \ \land (v_1,v_2,v_3) < (u_1,u_2,u_3)) \\
\qquad\qquad\quad\ \rar (\exists x(v_1(x)\land u_2(x)\land\neg v_2(x))\\
\qquad\qquad\qquad\quad\lor \exists y (v_2(y)\land u_1(y)\land\neg
v_1(y)))) .
\ea
\]

\bigskip\noindent
{\bf Remark:} Proposition~\ref{prop:2nd-order}$'$ tells that each of
the formulas in {\bf 1}, {\bf 2}, {\bf 3} are equivalent to each
other.

\bigskip\noindent
{\bf 4. First-Order Loop Formula for Sentence $F$ (Using $\fnes$) :}
Let $Y_1=\{p(z)\}$,$Y_2=\{q(z)\}$, $Y_3=\{r(z)\}$,
$Y_4=\{p(z),q(z)\}$. Set $\{Y_1,Y_2,Y_3,Y_4\}$ is a complete set of
loops.

Under the assumption~$F$,
\begin{itemize}
\item  $\flf_F(Y_1)$ is equivalent to the universal closure of
\[
\ba l
 p(z)\rar\neg\forall xy ([q(x)\rar p(x)\land x\ne z] \\
\qquad\qquad\qquad\land  [p(y)\land y\ne z\rar q(y)] \\
\qquad\qquad\qquad\land  [\neg r(x)\rar p(x)\land x \ne z]). \ea
\]

\item  $\flf_F(Y_2)$ is equivalent to the universal closure of
\[
\ba l  q(z) \rar \neg \forall xy ([q(x)\land x \neq z\rar p(x)] \\
\qquad\qquad\qquad\land  [p(y)\rar q(y)\land y \neq z]).
\ea
\]

\item  $\flf_F(Y_3)$ is equivalent to the universal closure of
\[
r(z) \rar \bot.
\]

\item  $\flf_F(Y_4)$ is equivalent to the universal closure of
\[
\ba l p(z)\land q(z)\rar \\
\quad\neg\forall xy ([q(x)\land x \neq z \rar p(x) \land x \neq z]\\
\qquad\quad\land     [p(y) \land y \neq z \rar q(y)\land y \neq z] \\
\qquad\quad \land    [\neg r(x)\rar p(x)\land x \neq z]). \ea
\]
\end{itemize}

\bigskip\noindent
{\bf 5. First-Order Loop Formula for Nondisjunctive Program
 (Using $\fes$):} See Example~\ref{ex:1}.

\bigskip\noindent
{\bf 6. First-Order Loop Formula when $\Pi$ is understood as an
 extended program (Using $\i{EFES}$) : }
Consider the same $Y_i$ as before.

Under the assumption $\Pi$,

\begin{itemize}
\item  $\flf_{\Pi}(Y_1)$ is equivalent to the universal closure of
\[
\ba l p(z) \rar ( \exists x ( q(x) \land \neg (p(x) \land x \neq
z)) \\
\qquad\quad\lor\ \exists x (\neg r(x) \land \neg(p(x)\land x \neq
z))). \ea
\]

\item  $\flf_{\Pi}(Y_2)$ is equivalent to the universal closure of
\[
\ba l q(z) \rar \exists y (p(y) \land \neg (q(y)\land y \neq z)).
\ea
\]

\item  $\flf_{\Pi}(Y_3)$ is equivalent to the universal closure of
\[
\ba l r(z) \rar \bot. \ea
\]

\item  $\flf_{\Pi}(Y_4)$ is equivalent to the universal closure of
\[
\ba l (p(z) \land q(z)) \rar(\exists x ( (q(x)\land x \neq z) \land
\neg (p(x) \land x \neq z)
)\\
\qquad\qquad\qquad\quad\lor\ \exists y ((p(y) \land y \neq z) \land

\neg
(q(y)\land y \neq z)) \\
\qquad\qquad\qquad\quad\lor\ \exists x (\neg r(x) \land
\neg(p(x)\land x \neq z))). \ea
\]
\end{itemize}

\bigskip\noindent
{\bf Remark:} Proposition~\ref{prop:lf-extended-program} tells that the
sets of formulas in each of {\bf 4}, {\bf 5}, {\bf 6} are equivalent
to each other, under the assumption $F$. In view of
Proposition~\ref{prop:fcsl}, each set conjoined with $F$ is equivalent
to each of the formulas in {\bf 1}, {\bf 2}, {\bf 3}.

\end{document}